\documentclass[preprint,12pt,authoryear]{elsarticle}

\usepackage{amssymb}
\usepackage{amsmath,amssymb,amsfonts}
\usepackage{algorithmic}
\usepackage{graphicx}
\usepackage{textcomp}
\usepackage{academicons}
\usepackage{scalerel}
\usepackage{tikz}
\usepackage{bm}
\usepackage{fancyvrb,cprotect}
\usetikzlibrary{svg.path}
\usepackage{hyperref} 
\usepackage{adjustbox}
\usepackage[authoryear]{natbib}
\usepackage{xcolor}
\usepackage{tabularx} 
\usepackage{array}    
\usepackage{float}
\usepackage{ragged2e}
\usepackage[flushleft]{threeparttable}
\usepackage{multirow} 
\usepackage{booktabs} 
\usepackage{array} 

\makeatletter
\def\ps@pprintTitle{%
  \let\@oddhead\@empty
  \let\@evenhead\@empty
  \def\@oddfoot{\reset@font\hfil\thepage\hfil}
  \let\@evenfoot\@oddfoot
}
\makeatother

\journal{Engineering Applications of Artificial Intelligence}

\begin{document}

\begin{frontmatter}



\title{Enhancing Next Destination Prediction: A Novel LSTM Approach Using Real-World Airline Data}

\renewcommand{\thefootnote}{\fnsymbol{footnote}}
\author{Salih Salihoglu{$^{a*}$}}
\author{Gulser Koksal{$^{b}$}}
\author{Orhan Abar{$^{c}$}}

\affiliation{organization={Department of Industrial Engineering, Middle East Technical University},
            city={Ankara},
            country={TURKEY}}
\affiliation{organization={Department of Industrial Engineering, TED University},
            city={Ankara},
            country={TURKEY}}
\affiliation{organization={Department of Computer Engineering, Osmaniye Korkut Ata University},
            city={Osmaniye},
            country={TURKEY}}



\begin{abstract}

In the modern transportation industry, accurate prediction of travelers' next destinations brings multiple benefits to companies, such as customer satisfaction and targeted marketing. This study focuses on developing a precise model that captures the sequential patterns and dependencies in travel data, enabling accurate predictions of individual travelers' future destinations. To achieve this, a novel model architecture with a sliding window approach based on Long Short-Term Memory (LSTM) is proposed for destination prediction in the transportation industry. The experimental results highlight satisfactory performance and high scores achieved by the proposed model across different data sizes and performance metrics. This research contributes to advancing destination prediction methods, empowering companies to deliver personalized recommendations and optimize customer experiences in the dynamic travel landscape.

\end{abstract}

\begin{keyword}
next destination prediction, LSTM, deep learning
\end{keyword}

\end{frontmatter}

\section{Introduction}
\label{sec:introduction}
Within the global economic framework, the travel industry plays a pivotal role in promoting economic growth in addition to facilitating the movement of people and goods both domestically and internationally. The aviation industry is one of the major players in this field. In this dynamic industry, the problem of next destination prediction is of utmost importance. Predicting a person's next destination using their past travel behavior is a big challenge. Precise predictions of destinations enable the delivery of customized services and focused marketing, which result in a more individualized and pertinent consumer experience. Moreover, it enhances the effectiveness of travel logistics. Consequently, this predictive strategy is vital in refining operational productivity and boosting overall consumer satisfaction within the travel industry.

Next destination prediction is challenging because it requires capturing the complex patterns and dependencies in the previous travel history of the individual. Traditional machine learning techniques, such as decision trees and logistic regression, have limited ability to capture the temporal dependencies and nonlinear patterns in the trajectory data. On the other hand, deep learning methods have demonstrated promising results in the next destination prediction. Recurrent Neural Networks (RNNs) and their variants such as Long Short-Term Memory (LSTM) are among the most effective methods, which can capture long-term dependencies and deal with more complex problems \citep{yu2019review}. However, they are not without limitations. For instance, LSTMs, while effective in handling long-term dependencies, can be computationally intensive and may struggle with extremely large datasets or when integrating diverse contextual features like time of day, weather conditions, and traffic congestion. These challenges highlight the need for further refinement and development of more efficient models that can leverage the strengths of LSTM while addressing its shortcomings.

The primary objective of this study is to examine and develop an efficient prediction model and methodology to address the next destination prediction problem. This research assesses the performance of the proposed prediction model and measures its effectiveness in terms of accuracy and scalability. The proposed method in this study utilizes a unique real-world airline dataset with limited features, and its applicability extends beyond the airline industry to other transportation industries as well. The task at hand involves solving a multiclass classification problem, where the target variable comprises the cities served by the airline company.

This study makes a valuable contribution to the existing literature by examining the proposed approach's strengths and weaknesses and shedding light on its effectiveness in addressing the prediction problem. One notable contribution lies in the uniqueness of the dataset employed. It is worth emphasizing that the dataset used in this study differs from those previously investigated in the literature. Additionally, depending on the dataset's characteristics, various feature engineering techniques are employed to tailor the methods accordingly. Multiple factors are considered to capture and model customer behavior effectively. Importantly, these methods are applicable not only to the airline industry but also to diverse sectors within the transportation industry.
\section{Related Work}

\label{sec:relatedwork}

This review synthesizes recent developments in next destination prediction, emphasizing the use of RNNs and their subclasses, particularly LSTM networks. RNNs, known for their sequential data processing capabilities, have been applied in various contexts, including word embedding for sentence modeling (\citealp{mikolov2010recurrent, mikolov2011extensions, mikolov2011rnnlm}), sequential click prediction \citep{zhang2014sequential}.

Spatial Temporal-RNNs \citep{liu2016predicting} and RNNs with suprisal-driven zoneout \citep{zhang2018predicting} illustrate advancements in handling spatial-temporal contexts and enhancing robustness and training efficiency. \citet{rossi2019modelling} demonstrate an effective RNN approach for taxi journey predictions using Location-Based Social Networks data, highlighting the method's suitability in geographically diverse cities.

LSTM models, overcoming the vanishing gradient problem of traditional RNNs, are increasingly used for sequence modeling tasks, including destination prediction. Their ability to manage long-term dependencies and selectively retain information makes them ideal for modeling sequential data. While studies by \citet{ishihara2021weighted}, \citet{lu2019destination}, and \citet{kim2017probabilistic} have demonstrated significant advancements in the application of various LSTM models for predicting destinations and movement trajectories, there remains room for enhancement. This potential for advancement is particularly relevant in addressing specific challenges such as handling larger datasets, integrating more complex contextual features, and improving computational efficiency. Recognizing these opportunities for improvement, this study aims to advance LSTM-based methodologies to achieve even more accurate and efficient next destination predictions.

In the realm of destination prediction, diverse methodologies have been explored with varying degrees of success. For instance, \citet{jiang2021dp} develop a Bayesian personalized ranking model called DP-BPR, achieving a 78\% accuracy with the top 5 recommendations. This model innovatively integrates user, time, and location embeddings to maximize the probability of predicting a user’s next destination. Similarly, \citet{yang2015destination} introduce the PROFILE method, utilizing taxis' GPS trajectories in Beijing to predict destinations with improved run time efficiency and accuracy compared to the SubSynEA method. Their approach clusters trajectories based on key features, demonstrating the utility of discriminative methods in destination prediction.

\citet{dai2018cluster} adopt a cluster-based methodology using various machine learning algorithms for the Citi Bike system in New York City. Their approach, which includes Station-level Clustering, Geographic Clustering, and Compound Station Clustering, resulted in an average prediction accuracy of 39.3\% using Random Forests with their Compound Station Clustering method. This study underscores the potential of clustering methods in handling large-scale urban mobility data.

Moreover, \citet{mathew2012predicting} employ Hidden Markov Models to predict future positions based on historical location data, achieving a 26.4\% accuracy for the top-5 most probable locations. Their approach effectively captures patterns in location histories but also reveals limitations in prediction accuracy.

\citet{zong2019trip} develop a model using multi-day GPS data, applying a combination of Markov chains and Multinomial logit models for pre-trip and during-trip destination predictions. Their methodology, which considers weekdays and weekends separately, showed high accuracies of up to 91.04\% for weekday predictions, but a notable decrease in accuracy during weekends.

While these studies represent significant advancements in destination prediction, they also highlight the challenges of achieving high accuracy and consistency under varying conditions and datasets. This emphasizes the need for robust, flexible models that can handle complex real-world data. 

The review also discusses route prediction techniques such as Bayes classifiers, histogram matching, and probabilistic models, which are similar to destination prediction even though they are different (\citealp{marmasse2002user,patterson2003inferring,krumm2013destination}).

A wide range of approaches to next destination prediction have been documented in the literature. One of these approaches is LSTM, which is a successful model because of its sophisticated processing of sequential data and long-term dependencies. This study contributes to this expanding field by offering a novel LSTM-based model designed specifically for the requirements of the travel industry.
\section{Methodology}

\label{sec:methodology}

In this section, the proposed method for predicting the next destination is provided. This problem is addressed by the introduction of a novel model architecture that combines LSTM and a sliding window technique. This novel method predicts customers' next destinations with accuracy by taking advantage of the sequential patterns and dependencies found in travel data. It provides a comprehensive framework to enhance destination prediction accuracy by accounting for the continuous flow of trips and capturing the temporal relationships between destinations. The architecture, underlying algorithms, and specifics of the proposed approach are covered in detail.

Consumers' travel histories reveal unique patterns and trends that can be used to gather useful information about their preferences and behaviors. Researchers are able to predict customers' next destinations with greater accuracy when these patterns are accurately captured and analyzed. Relying on past travel records, the problem is to predict the next destination, where destination order matters. Travel data may be categorized as sequential data since the itineraries of the trips are recorded in that order. That means a customer's previous trip sequence contains useful information that can be used to forecast where they will travel to next. This task proves to be well suited for LSTM because of its capacity to capture patterns and temporal dependencies within destination sequences. It enables comprehension of the connections between several prior destinations and the current one by modeling long-term dependencies effectively. To further accommodate customers with varying travel histories, LSTM can also handle variable-length sequences. LSTM is capable of personalizing predictions by adding context and extra features like seasonality and user preferences.The primary objective of employing this model is to evaluate the efficacy of an advanced deep learning architecture in accurately predicting the next destinations of customers.

The authors suggest a sliding window approach as a key component of the LSTM model to handle the next destination problem more successfully. This method systematically uses sequential trip pairs within a dynamic window. Overall, the proposed method makes a number of important advances. First, it involves using the sliding window method in a customized way that is intended for the prediction of the next destination. A thorough examination of the sequential nature of the data is made possible by the sliding window, which is used by combining trip pairs to create windows. Secondly, the assumption of continuous trips and their sequential arrangement as chains ensures the incorporation of temporal dependencies and patterns into the model. Lastly, a novel model architecture is developed, integrating sliding windows to effectively model both short-term and long-term dependencies within the trip sequence. Collectively, these contributions enhance the understanding and accuracy of next destination prediction by leveraging the advantages of the sliding window approach within the LSTM framework.

\subsection{Problem Definition}
\label{chp:problem definition}

The next destination prediction problem refers to the task of predicting where an individual is likely to go or travel to next, based on available information about their previous movements, preferences, context, and other relevant factors. Given a dataset consisting of $m$ customers' travel histories, the objective is to predict the next destination a customer will likely visit in their future travel. The prediction task involves assigning a single destination label to each customer based on their past travel attributes.

Let's denote that there is a set of customers $C = \{c_1, c_2, ..., c_m\}$, who has previously traveled with the transportation company, where $m$ represents the total number of customers. For each customer, denoted as $c_i$, a sequence of historical trip records is available. This sequence, represented as $T_i = [(o_1, d_1), (o_2, d_2), ..., (o_n, d_n)]$, captures the customer's past travel experiences. Here, $n$ represents the number of trips in the sequence, while $o_k$ and $d_k$ denote the origin and destination cities of the $k^{th}$ trip, respectively. Let $\bold{F_i}$ represent the additional features vector for customer $c_i$, which can be expressed as $\bold{F_i} = \{\bold{f_1}, \bold{f_2}, \ldots, \bold{f_n}\}$. Here, $\bold{f_k}$ denotes the vector of additional features for the $k^{th}$ trip, including attributes such as the weekday of the trip, journey type, and so on. The length of each vector $\bold{f}$ is denoted by $s$, which may vary depending on the dataset.

$Problem:$ Given the current (or the most recent recorded) location, $o_{n+1}$, of a customer $c_i$, the historical trip records $T_i$ and additional features vector $\bold{F_i}$, predict the next destination city, $d_{n+1}$, for that customer $c_i$.

$Assumptions:$ Firstly, the prediction is based solely on the available information up to the current time, without any knowledge of future trip records. Secondly, the prediction is limited to the cities served by the transportation company. The model focuses on destinations within the company's service area and does not consider locations outside of this scope. Thirdly, it is assumed that there is enough data to observe patterns. Sufficient historical travel data are available for customers, enabling the identification of patterns, dependencies, and relationships within the data. Lastly, it is assumed that the current location of the customer is known. The transportation company has access to up-to-date information about the customer's current location, which is essential for predicting their next destination accurately. These assumptions guide the development of the prediction model and help ensure that it effectively utilizes available data and current location information to make accurate predictions within the transportation company's service area.

Transportation companies can employ various methods to determine the current location of customers and predict their next destination. These methods include ticketing or reservation systems where customers provide their current location details during the booking process. Mobile applications or account information can also provide location data collected from customers' devices or accounts. Loyalty programs may require location information for membership, while check-in systems or beacons at stations enable tracking of customer locations within the transportation network. Wi-Fi or Bluetooth connectivity on vehicles or at stations can be utilized to gather customer presence and location data. Additionally, partnerships with mobile network operators can provide access to aggregated location data from mobile devices. Also, the most frequent departure location of customers may be available in the dataset, as is the case in the dataset used in this study. This most frequent departure location can be used as the current location. Also, transportation companies can accurately determine the current location of customers by using their sources. Companies need to take great care when handling location data from customers in order to respect customer privacy and comply with data protection laws.

This problem in this study can be formulated as a multiclass classification problem, where the set of possible destination labels forms the classes. Each customer's historical travel attributes serve as the input features. In this context, the time intervals between trips can be in various temporal granularities, such as a day, week, month, or year. To formulate this problem as a multiclass classification task, the set of possible destination labels can be defined as $D = \{d_1, d_2, ..., d_k, ..., d_p\}$, where $p$ represents the total number of distinct cities served by the transportation company. Each destination city $d_k$ represents a class label. The goal is to learn a predictive model that can accurately classify the next destination label $d_{n+1}$ for each customer $c_i$ based on their historical travel attributes. The model takes the input features, including the current location, $o_{n+1}$, the historical trip records $T_i$ and the additional features vector $\bold{F_i}$, and maps them to the corresponding destination label $d_{n+1}$ from set $D$.

The primary aim of this study is to embark upon the development of a robust and accurate predictive model specifically designed to predict the next destinations of customers. The overarching goal is to maximize the number of correctly classified next destinations. In the context of this study, the target variable of interest comprises the cities to which customers travel. This research seeks to enhance the precision and reliability of the developed model.

The study must consider several factors that influence destination predictions. One important aspect is the identification of patterns specific to individual customers, which can aid in making accurate predictions. For instance, a customer who consistently travels to Europe during the summer season is likely to continue this pattern in the future. Therefore, it is crucial to thoroughly analyze each customer's travel history to uncover and understand such recurring patterns. Additionally, seasonality plays a vital role in travel patterns, as certain destinations experience increased popularity during specific seasons, such as summer holidays. Taking seasonality into account improves the precision of destination predictions. Furthermore, it is essential to consider the impact of business-related travel. Individuals who frequently travel to specific destinations for business purposes are more likely to visit those places again in the near future. Identifying such customers and their associated travel patterns can significantly enhance the accuracy of predictions.

\subsection{Proposed Method}
\label{proposedmethod}

Given a dataset D consisting of customer trip sequences, where each sequence $T_i$ represents the historical trips of customer $c_i$ as mentioned in Section \ref{chp:problem definition}, it is aimed to train the model to accurately predict the next destination.

For this purpose, a sliding window approach is proposed. In general, such approaches are commonly used in tasks involving sequential data analysis and prediction. They are based on windows moving through a sequence of data, capturing subsets of data at each step. Utilizing the sliding window technique, the travel data is segmented into overlapping fixed-length windows. By using a series of prior destinations as inputs for predicting the subsequent destination, this technique enables the LSTM model to identify temporal patterns and dependencies within the data. In this regard, the sliding window method helps to uncover both short-term and long-term patterns in the consumer's travel history.

Regarding short-term dependencies, the sliding window method enables the LSTM model to recognize recent travel behaviors within a predetermined window of past trips. Focusing on a limited selection of the latest trips allows the model to concentrate on the customer's immediate actions and preferences. This is beneficial for identifying short-term variations and tendencies in the customer's travel decisions. In the case of long-term dependencies, with the help of sliding window method, the model can accommodate a variety of historical trips, including more distant and recent ones, by swiping the window along the trip sequence. As a result, the customer's enduring preferences and tendencies can be captured by the model. Over time, the model can learn and comprehend the customer's destination choices which are influenced by broader travel habits and patterns by integrating a wider window of historical data.

Formally, let $T_i=[(o_1,d_1),(o_2,d_2),...,(o_n,d_n)]$ be the sequence of trips for customer $c_i$ and $n$ is the number of trips. A sliding window of size $w$ is defined over the sequence of $T_i$ as:

\begin{equation}
\label{eqn:slidwin}
\begin{split}
\bold{W}_i = & \{[(o_1,d_1),(o_2,d_2),...,(o_w,d_w)]_i^1, \\
& [(o_2,d_2),(o_3,d_3),...,(o_{w+1},d_{w+1})]_i^2,..., \\
& [(o_{n-w},d_{n-w}),(o_{n-w+1},d_{n-w+1}),...,(o_{n-1},d_{n-1})]_i^{n-w} \}
\end{split}
\end{equation}

The dataset with custom features for customer $c_i$ can be denoted as:

\begin{equation}
\label{eqn:entry}
\begin{split}
\bold{R}_i = & \{(\bold{w}_i^1,\bold{e}_i^1),(\bold{w}_i^2,\bold{e}_i^2),...,(\bold{w}_i^{n-w},\bold{e}_i^{n-w})\} \\
= & \{([(o_1,d_1),(o_2,d_2),...,(o_w,d_w)]_i^1,\bold{e}_i^1), \\
& ([(o_2,d_2),(o_3,d_3),...,(o_{w+1},d_{w+1})]_i^2,\bold{e}_i^2),..., \\
& ([(o_{n-w},d_{n-w}),(o_{n-w+1},d_{n-w+1}),...,(o_{n-1},d_{n-1})]_i^{n-w},\bold{e}_i^{n-w}) \}
\end{split}
\end{equation}

\bigskip

Each tuple in $\bold{R}_i$ for customer $c_i$ represents an entry, where $n-w$ is the number of windows. The feature vector $\bold{E_i} = \{\bold{e}_i^1, \bold{e}_i^2, \ldots, \bold{e}_i^{n-w}\}$ includes custom features for each window that are derived from additional features vector $\bold{F_i}$, defined in Section \ref{chp:problem definition}, present in the original dataset. These custom features are created based on the dataset and are subject to the feature engineering choices made by the researcher. The custom features can be modified or adjusted as needed. If there are any additional features available, they can also be added to the feature vector $\bold{E}$. The set $\bold{R}_i$ is constructed in a way that consists of the historical trips of customer $c_i$. After splitting the dataset, the training set is formed by including all trips of the customer except for the last trip. The last trip is reserved for the test set. In the training process, the origin city of the last trip following the window in the training set is added as a feature before the final layer, and the model is trained to predict the destination city of this last trip associated with that window. The same process is applied to the test set, with the only difference being that the test set includes only the last trip among all the trips made by each customer. This approach ensures that the model is trained on the customer's historical travel data, excluding the last trip, and is then evaluated on its ability to predict the destination city of the last trip in the test set.

It is assumed that there may exist intermediary trips between the destination city of a trip and the origin city of the next trip, regardless of whether they are the same or different. These intermediary trips, if present, may involve modes of transportation not provided by the transportation company. The model focuses solely on predicting the next destination based on the available data and does not consider the specific means by which the customer reaches the origin city of the next trip after the destination city of the current trip.

\begin{equation}
\label{eqn:lossfunction}
M = argmin_M \sum_{i={1}}^{m} \sum_{j={1}}^{n-w} L(M(\bold{R}_i^j,o_i^{w+j}),d_i^{w+j})
\end{equation}

In Equation \ref{eqn:lossfunction}, $M(\bold{R}_i^j,o_i^{w+j})$ represents the trained model using historical data, the origin city of the trip following the window, and custom features for customer $c_i$ at entry $j$, where $j$ includes the window and related custom features with that window as represented as in Equation \ref{eqn:entry}. $m$ represents the number of customers in the dataset. In the training process, the origin city of the trip following the window, denoted as $o_i^{w+j}$, is provided to the model just before the final layer. Additionally, $d_i^{w+j}$ represents the true next destination associated with that window. The function $L$ represents a suitable loss function that measures the discrepancy between the predicted and true destinations. The goal is to find the function $M$ that minimizes this prediction error across all customers during the training stage. The performance of the model will be evaluated using the top-N F1 score metric in the test set. The following parts below will provide a detailed explanation of the entire process.

\begin{figure}[H]
    \center
    \includegraphics[scale=0.6]{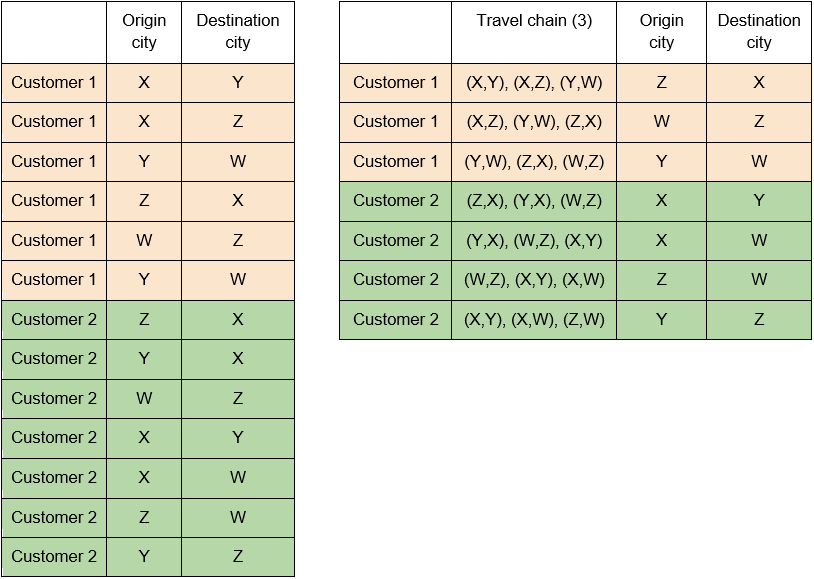}
    \caption{A Windowing Sample}
    \label{fgr:windowingsample}
\end{figure}

The first step of the model is converting the dataset into a window in which the past sequence is used as a feature in the LSTM model as in Equation \ref{eqn:slidwin}. For the specified window size, a group of trips is used to create a travel data chain. The travel data chain has the past trips of the customer with time-based orders. Other than the travel data chain, a set of features can also be created to contribute to the model’s prediction power as in Equation \ref{eqn:entry}. The windowing process is represented below with a window size of 3 (Figure \ref{fgr:windowingsample}). This is a parameter to prepare the dataset and different window sizes can affect the performance. In Section \ref{sec:experiments}, different window sizes will be tried to find the optimum performance.

After the windowing operation is applied, the dataset is converted into a sequential format that allows it to be easily integrated with an LSTM model. After the data is arranged in this sequential manner, it opens up to the temporal patterns and dependencies that are present throughout the dataset.

In addition, as Table \ref{tab:feature-explanation} illustrates, custom features can be developed using the existing features of the dataset to improve the model's predictive power. These unique characteristics are intended to catch extra details and subtleties that could lead to a higher degree of prediction accuracy. After the creation of the dataset, several preprocessing steps are implemented to ensure its suitability for feeding into the LSTM model. The LSTM model requires four distinct components as input: numerical features, categorical features, date features, and embedding features.

Starting with the numerical features, a numerical transformer pipeline is applied to handle missing values and standardize the features. To address missing values, a simple imputer is employed, which fills the gaps with the median values of the respective features. This imputation strategy ensures the preservation of data integrity and minimizes the impact of missing information. After that, the numerical features are normalized using a standard scaler to ensure consistency and prevent any one feature dominating the learning process.

The next step involves the categorical features, which are converted into a numerical format using a one-hot encoding technique. By means of this conversion, the LSTM model can efficiently comprehend and employ categorical data for both training and prediction purposes. Categorical variables are represented as binary vectors.

Finally, the features of embedding are discussed. To transform specific features into embedding vectors, an embedding layer is added before the LSTM model. By using this method, the sparsity issue related to high-dimensional sparse features is mitigated. One-hot encoding is sufficient for categorical features with few categories, but embeddings are preferred for features with many categories. By representing these features as lower-dimensional embedding vectors, the model benefits from more efficient representation and enhanced generalization capabilities.

Following the preprocessing of the features and the incorporation of embedding layers for the city-based features, the next step involves utilizing a concatenation layer to merge all the input components. This concatenated layer serves as the input for the subsequent model architecture. In this specific case, two LSTM layers are employed to capture the sequential dependencies and temporal patterns within the data effectively. Furthermore, an embedding layer specifically designed for the origin city feature is added to the model architecture. 

\begin{figure}[H]
    \center
    \includegraphics[scale=0.4]{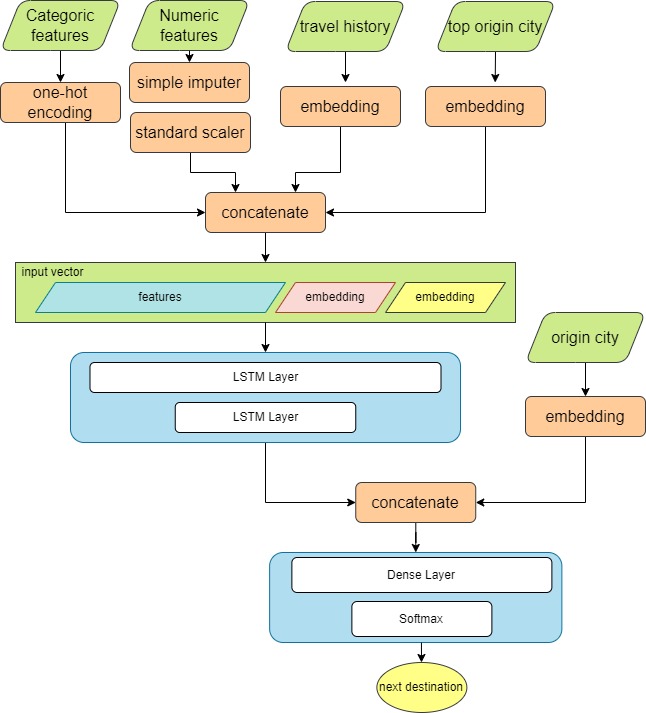}
    \caption{The Diagram of the Proposed Model}
    \label{fgr:diagramlstm}
\end{figure}

As the last layer before the output, a dense layer is also added to the model. This dense layer is responsible for transforming the learned representations into appropriate output predictions. Using a softmax activation function, the raw outputs are transformed into a probability distribution over the various categories that make up the model's desired output. To provide a visual representation of the model architecture, Figure \ref{fgr:diagramlstm} presents a diagram depicting the various components and connections within the model.

\bigskip
\section{Application of the Proposed Method and Results}
\label{sec:experiments}
In this chapter, the proposed method is applied on a real-world dataset obtained from an airline company. The chapter presents and analyzes the experimental results, with a primary focus on assessing the performance of the proposed method in predicting the next destination of customers. The dataset serves as the basis for evaluating the performance and efficiency of the proposed approach. To gain deeper insights, several experiments are conducted to investigate the impact of different parameters on the performance of the proposed method. The results obtained from these experiments are presented in detail, and the implications of these results are thoroughly discussed.

\subsection{Characteristics of Data}
\label{Characteristics of Data}

This study utilizes a dataset obtained from an airline company. It contains comprehensive information about customers who have traveled by air in the past five years. The dataset includes a large volume of flight-related records, with 115 million entries. Within this dataset, there are 36 million unique customers, showcasing the diversity of individuals represented. Among the customers, approximately 20 million have taken only one flight, while 16 million have a history of multiple flights. To prepare the dataset for modeling, a selection process is employed. Customers with more than 17 flights including the last trips of training and test sets after creating windows are considered, aligning with the maximum window size of 15. The dataset is further refined by focusing on the top 16 cities, which account for about 90 percent of the flights. This selective approach enables targeted analysis and prediction of the most significant and frequently visited destinations. As a result of these procedures, the final dataset consists of 222,000 distinct customers, forming the basis for subsequent analyses. Subsets of this cohort are strategically chosen to facilitate specialized investigations and tailored analyses within specific customer segments.

Features are as follows:

\begin{table}[H]
\centering
\caption{Features in the Dataset}
\label{tab:dataset-columns}
\begin{tabular}{lll}
\hline
Column name      & Explanation                         \\ \hline
PRIM\_TKT\_NUM   & Customer ticket number                \\
CITY\_NM         & Arrival city                           \\
CNTRY\_NM        & Arrival country                          \\
ORG\_CITY\_NM    & Departure city                          \\
ORG\_CNTRY\_NM   & Departure country                      \\
CUST\_KEY        & Customer key                            \\
SEG\_LCL\_DEP\_DT& GMT departure time                \\
ORGN\_AP         & Departure airport code                     \\
DSTN\_AP         & Arrival airport code                       \\
DOM\_INTNL\_FLAG & Flight type domestic or international  \\
JRNY\_TYP        & Journey type                    \\
TOP1\_ORG\_MKT\_REGN & The region that the customer departs most  \\
TOP1\_ORG\_CTY   & The city that the customer departs most  \\
\hline
\end{tabular}
\end{table}

The original dataset contains the aforementioned features, yet not all of them are utilized in the subsequent analysis. Alongside the existing features, additional custom features are created to enhance the prediction capabilities of the model. These custom features are specifically engineered to capture additional insights and patterns that can contribute to improved predictive performance.

In the following sections, these custom features will be described in detail, outlining their purpose and the rationale behind their inclusion. By incorporating these custom features, it is aimed to augment the richness of information available to the model, enabling it to uncover more intricate relationships and dependencies within the data.

\subsection{Creating the Datasets}
\label{overall_alg}

The first step before the application of the models is data preprocessing. In this step, data cleaning, transformation, integration, reduction, and sampling are applied to transform raw data into a format that can be easily analyzed and interpreted. In the original dataset, a significant number of erroneously recorded city names are identified and subsequently rectified through a rigorous analysis process. To enhance memory utilization during the modeling phase, variable types are adjusted to accommodate the substantial size of the dataset, which demands a substantial memory allocation. To ensure data integrity, rows with identical values in the arrival and departure city fields are excluded from the analysis. Furthermore, rows containing null values in either the arrival city, origin city, or top origin city columns are also eliminated from the dataset, ensuring the reliability and quality of the subsequent analysis.

The following steps are made for the LSTM model. Because of the nature of the problem, the dataset should be derived from different customers to generalize the behavior pattern and should contain all the flights of each customer for the sake of completeness. In this regard, a group of customers with different sizes is chosen as the subset. The random sampling method is applied to choose a subset since this is a customer-based model. Three different levels of sizes are chosen to test the effect of the customer size on the results. 5000, 15000, and 25000 are chosen as the low, medium, and high levels. All the flight data of these customers are gathered to create a dataset. 

Due to the sequential character of the dataset, the flights should be in a time-based order. Also, to validate the model properly, the test set should be chosen with the latest flights of each customer, as indicated in Figure \ref{fgr:windowingsample}. So, the last flight of each customer is taken as the test set and the remaining is used to create the training set by using the window size parameter. As explained in Figure \ref{fgr:windowingsample}, the window size is an important parameter for creating a dataset since it directly affects the dimensionality of the feature set and so the performance of the model. Three different window sizes are selected to test the effect of the window size on the performance. 5, 10, and 15 are the three levels and the dataset is created with these window sizes for each customer group. Also, since the last flight of the customer is used in the test set, a k-fold cross-validation strategy cannot be employed. Since there is enough data to create more datasets, for each customer size, 5 different customer sets are selected to be used as validation. 

\begin{figure}[H]
    \center
    \includegraphics[scale=0.7]{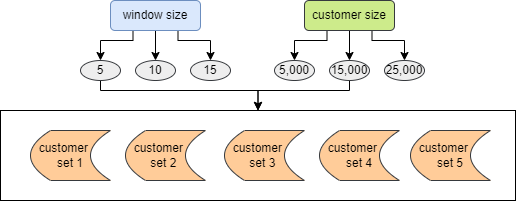}
    \caption{Data Splitting for Validation}
    \label{fgr:datasplit}
\end{figure}
    
In Figure \ref{fgr:datasplit}, the selection process of customer datasets is demonstrated. For each combination, different customer sets are used in the model and the average scores of these five different customer sets are taken as the final score for each parameter setting.

As a summary, a total of 45 datasets are created for 3 different customer sizes and 3 window sizes, in which each different parameter setting has 5 different datasets. The summary table for the details and shapes of the datasets is given below (Table \ref{datasetsummarylstm}).

\begin{table}[H]
\centering
\caption{Dataset Summary Table}
\label{datasetsummarylstm}
\begin{tabular}{>{\centering\arraybackslash}p{2cm} >{\centering\arraybackslash}p{2cm} >{\centering\arraybackslash}p{3cm} >{\centering\arraybackslash}p{3cm}}
\toprule
Customer size & Window size & Average number of rows in train data & Number of rows in test data \\
\midrule
5K  & 5  & 124K & 5K \\
5K  & 10 & 110K & 5K \\
5K  & 15 & 95K  & 5K \\
15K & 5  & 378K & 15K \\
15K & 10 & 333K & 15K \\
15K & 15 & 289K & 15K \\
25K & 5  & 626K & 25K \\
25K & 10 & 552K & 25K \\
25K & 15 & 477K & 25K \\
\bottomrule
\end{tabular}
\end{table}

\subsection{Selection of the Performance Metric}
\label{performancemetric}

The choice of performance metric for the next destination prediction depends on the specific requirements and objectives of the task. While the F1 score is commonly used for binary classification tasks, it may not directly address the top-N prediction scenarios where the goal is to predict the most likely destinations among a larger set of options. In such cases, the top-N F1 score can be a suitable metric. It is widely recognized as a popular evaluation metric for ranking tasks \citep{liu2016predicting}. There are several reasons and advantages to using this metric. Next destination prediction often involves ranking a set of potential destinations based on their likelihood. The top-N F1 score evaluates the model's ability to rank the correct destination within the top-N predictions. It evaluates the model's top-N predictions in terms of accuracy and recall by taking into account both false positives and false negatives within that subset. In real-world scenarios, users are typically interested in a limited number of most probable destinations. 

The top-N F1 score provides a more practical evaluation by considering the model's performance in identifying the relevant destinations within the top-N predictions. It focuses on the accuracy of the most important predictions rather than evaluating the entire prediction space. Since there are frequently a lot of possible destinations in the next destination prediction, it is not practical to assess the model's performance on each class separately. This approach reduces computational complexity while still capturing the model's ability to identify the most relevant options. The top-N F1 score aligns with the real-world scenario where users are typically presented with a limited number of destination options. For all the specified reasons, the top-N F1 score is utilized in this study.

\subsection{Experimental Runs}

LSTM requires a great amount of computational power to run \citep{hossain2023recurrent}. They can be computationally intensive, especially if the network is large or the training dataset is very large. High-power CPUs and GPUs are used to run LSTM models on big datasets. The dataset in this study has 115 million rows of flight data for over 36 million customers which can be considered quite big. Because of the limited time and resources, a subset-choosing strategy is employed to run the proposed method and obtain results for analyses.

\subsection{Results and Discussion}

In this section, the experimental results of the proposed model are presented and discussed. As previously elucidated in Section \ref{performancemetric}, the chosen performance metric is the top-N F1 score, which is defined to take into account the precision and recall of the model's predictions for the top-N most probable classes. The rationale behind utilizing the top-N F1 score is its ability to prioritize the identification of the most likely classes, rather than solely focusing on the top 1 prediction. The tables display the top-1, top-3, top-5, and top-7 F1 scores for both models. By evaluating the model's performance across these different prediction scenarios, a comprehensive understanding of its predictive capabilities can be attained. Following the model specifications detailed in Section \ref{proposedmethod}, the subsequent steps are meticulously executed to ensure a robust implementation that maximizes the potential of the LSTM architecture for accurate prediction.

The initial step of the model involves transforming the dataset into a window, where the preceding sequence is employed as a feature within the LSTM model. Within the defined window size, a collection of flights is selected to form a flight chain, encompassing the customer's previous flights in chronological order. In addition to the flight chain, a set of features is generated to enhance the predictive capability of the model. Various window sizes will be experimented with in order to identify the optimal performance. Following the implementation of the windowing operation, the data undergoes conversion into a sequential format, rendering it suitable for input into an LSTM model. Apart from the customer key, the dataset comprises additional columns including the origin city, destination, flight date, top origin city for each customer, flight type as a domestic or international flight, and journey type as one-way or return.

Furthermore, in the feature engineering step, certain features are crafted to enhance the predictive capabilities of the model (Table \ref{tab:feature-explanation}). Some of these features are derived from existing ones, while others are specifically created for each flight chain. For each flight, the day, week, month, and weekday of the flight date are extracted. The season of the flight is determined based on the month of the flight date. These extracted features will be employed in the derivation of additional window-specific features. The time difference between the first and last flights within a window is computed as the duration of the window. Additionally, the average flight day within a window is calculated to represent the frequency of travel for a given customer. The total count of domestic flights and the number of return flights within a window are also captured.

\begin{table}[ht]
\centering
\caption{Custom Features}
\label{tab:feature-explanation}
\begin{tabular}{>{\arraybackslash}p{5cm} >{\arraybackslash}p{7cm}}
\toprule
\textbf{Feature name} & \textbf{Explanation} \\
\midrule
Average day difference & Day difference between the first and the last flight of the window \\
Domestic flight count & Number of domestic flights \\
Return trip count & Number of return flights \\
First season & The season of the first flight \\
Last season & The season of the last flight \\
Days & Day of the flight date \\
Months & Month of the flight date \\
Weekdays & Weekday of the flight day \\
Flights array & Past flights of the customer in chronological order \\
\bottomrule
\end{tabular}
\end{table}
Once the dataset is created, several preprocessing steps are implemented to ensure its compatibility with the LSTM model, as described in Section \ref{proposedmethod}. The LSTM model requires four distinct components as input: numerical features, categorical features, date features, and embedding features. Therefore, appropriate preprocessing techniques are applied to each component to properly prepare the data for feeding into the LSTM model.

In this scenario, the city variable exhibits numerous alternative values, and employing one-hot encoding would lead to a high-dimensional and sparse feature matrix. To address this issue, embeddings are employed to represent these features more efficiently. Consequently, an embedding layer is utilized for the flight chain, top origin city, and origin city. The flight chain and top origin city are provided as inputs to the architecture, while the origin city is specifically given at the final layer, just prior to the softmax function. This design choice is motivated by the need to train the model exclusively with the features, ensuring that the origin of the last flight does not influence the model parameters.

Numerical features: Average day difference, domestic flight count, return trip count

Categorical features: First season, last season, days, months, weekdays

Embedding features: Flights array, top origin city, origin city

Following the preprocessing of the features and the application of embedding layers to the city-based features, a concatenation layer is employed to combine all inputs. This concatenated layer serves as the input for the subsequent model, which consists of two LSTM layers. The first LSTM layer consists of 100 nodes, while the second layer is composed of 20 nodes. These specific values are selected after experimenting with different configurations. It is observed that using lower values adversely affects the model's performance, whereas higher values do not lead to any significant improvement in performance. Therefore, the chosen configuration of 100 nodes in the first layer and 20 nodes in the second layer strikes a balance that optimizes the performance of the LSTM model. Additionally, an embedding layer for the origin city is incorporated, followed by the inclusion of a dense layer. The output of the dense layer is then passed through a softmax function to convert it into categorical probabilities. Given that the model produces output in the form of multiclass categories, a softmax function is utilized for this purpose.

\begin{table}[H]
\centering
\caption{Results of the Proposed Model}
\label{resultsofthemodel}
\begin{tabular}{>{\centering\arraybackslash}p{1.5cm}>{\centering\arraybackslash}p{1.5cm}>{\centering\arraybackslash}p{1.5cm}>{\centering\arraybackslash}p{1.5cm}>{\centering\arraybackslash}p{1.5cm}>{\centering\arraybackslash}p{1.5cm}}
\toprule
Customer size & Window size & Top-1 F1 score & Top-3 F1 score & Top-5 F1 score & Top-7 F1 score \\
\midrule
5,000 & 5  & 0.73 & 0.85 & 0.91 & 0.94 \\
5,000 & 10 & 0.71 & 0.85 & 0.92 & 0.93 \\
5,000 & 15 & 0.72 & 0.85 & 0.91 & 0.93 \\
15,000 & 5 & 0.74 & 0.85 & 0.92 & 0.96 \\
15,000 & 10 & 0.74 & 0.86 & 0.92 & 0.95 \\
15,000 & 15 & 0.73 & 0.86 & 0.93 & 0.95 \\
25,000 & 5 & 0.76 & 0.88 & 0.93 & 0.96 \\
25,000 & 10 & 0.78 & 0.88 & 0.94 & 0.96 \\
25,000 & 15 & 0.78 & 0.89 & 0.93 & 0.97 \\
\bottomrule
\end{tabular}
\end{table}

In this model, two parameters used are customer size and window size as can be seen in Table \ref{resultsofthemodel}. Based on analysis of variance (ANOVA) of results, the main and interaction effects of customer size (CS) and window size (WS)  are calculated for four F1 scores (Top 1, Top 3, Top 5, Top 7), and the results obtained are provided in Table \ref{table:anova}.

It is evident from Table \ref{table:anova} that the only main factor CS exhibits statistically significant results for all evaluated F1 scores. The results for Top-1, Top-3, Top-5 and Top-7 F1 scores are $(p < .01, \eta^2= .893)$, $(p < .01, \eta^2= .966)$, $(p < .05, \eta^2= .500)$ and $(p < .01, \eta^2= .893)$, respectively.

\begin{table}[H]
\centering
\begin{threeparttable}
\caption{Results of Factorial ANOVA}
\label{table:anova}
\begin{tabular}{@{}lcccccc@{}}
\toprule
\textbf{F1 Score} & \textbf{Factor} & \textbf{Sum Sq} & \textbf{df} & \textbf{F} & \textbf{p} & \textbf{\(\eta^2\)} \\
\midrule
\multirow{4}{*}{Top 1} & CS          & 0.004876 & 1 & 45.308 & .001** & .893 \\
                       & WS          & 0.000012 & 1 & 0.111  & .753 & .000 \\
                       & CSxWS       & 0.000253 & 1 & 2.348  & .186 & .000 \\
                       & Error       & 0.000538 & 5 & --     & --   & --   \\
\midrule
\multirow{4}{*}{Top 3} & CS          & 0.001667 & 1 & 34.483 & .002** & .966 \\
                       & WS          & 0.000667 & 1 & 1.379  & .293 & .000 \\
                       & CSxWS       & 0.000250 & 1 & 0.517  & .504 & .000 \\
                       & Error       & 0.000242 & 5 & --     & --   & --   \\
\midrule
\multirow{4}{*}{Top 5} & CS          & 0.000600 & 1 & 16.364 & .010* & .500 \\
                       & WS          & 0.000017 & 1 & 0.455  & .530 & .000 \\
                       & CSxWS       & 0.000000 & 1 & 0.000  & 1.00 & .000 \\
                       & Error       & 0.000183 & 5 & --     & --   & --   \\
\midrule
\multirow{4}{*}{Top 7} & CS          & 0.001350 & 1 & 50.625 & .001** & .893 \\
                       & WS          & 0.000167 & 1 & 0.625  & .465 & .000 \\
                       & CSxWS       & 0.000100 & 1 & 3.750  & .111 & .000 \\
                       & Error       & 0.000133 & 5 & --     & --   & --   \\
\bottomrule
\end{tabular}
\begin{tablenotes}[para,flushleft]
\footnotesize
\item[*] \(p<.05\), ** \(p<.01\), CS: Customer Size, WS: Window Size
\item[] The p-value represents the likelihood of obtaining an outcome as extreme or more extreme than the one actually observed, under the assumption that the null hypothesis is correct \citep{moore1996basic}. Beyond the p-value from factorial ANOVA, reporting the effect size coefficient, which is sample size independent, is crucial for practical significance. Commonly used effect size coefficients in Factorial ANOVA results are $\eta^2$ and partial $\eta^2$ (\citealp{clark2012examination}; \citealp{norouzian2018eta}).
\end{tablenotes}
\end{threeparttable}
\end{table}

Following the results of the factorial ANOVA, further investigations have been conducted on CS that are significant and have medium to high effect sizes. In this context, the ‘TukeyHSD’ function in the ‘stats’ package \citep{team2010r} has been utilized to perform the Tukey multiple comparison test. The results obtained are provided in Table \ref{table:tukey_test}. As for WS, all results demonstrate no statistical significance.

\begin{table}[H]
\centering
\begin{threeparttable}
\caption{Post Hoc Comparisons of Customer Size using Tukey's Test}
\label{table:tukey_test}
\begin{tabular}{@{}llccccc@{}}
\toprule
\textbf{F1 Score} & \textbf{Contrast} & \textbf{Estimate} & \textbf{SE} & \textbf{df} & \textbf{t.ratio} & \textbf{p.value} \\
\midrule
\multirow{3}{*}{Top 1} & CS5K, CS15K & -0.0167 & 0.080943 & 4 & -1.768 & 0.2900 \\
                       & CS5K, CS25K & -0.0533 & 0.080943 & 4 & -5.657 & 0.0105* \\
                       & CS15K, CS25K & -0.0367 & 0.080943 & 4 & -3.889 & 0.0380* \\
\midrule
\multirow{3}{*}{Top 3} & CS5K, CS15K & -0.0667 & 0.080333 & 4 & -2.000 & 0.2276 \\
                       & CS5K, CS25K & -0.0333 & 0.080333 & 4 & -10.000 & 0.0013** \\
                       & CS15K, CS25K & -0.0267 & 0.080333 & 4 & -8.000 & 0.0029** \\
\midrule
\multirow{3}{*}{Top 5} & CS5K, CS15K & -0.01 & 0.08471 & 4 & -2.121 & 0.2006 \\
                       & CS5K, CS25K & -0.02 & 0.08471 & 4 & -4.243 & 0.0286* \\
                       & CS15K, CS25K & -0.01 & 0.08471 & 4 & -2.121 & 0.2006 \\
\midrule
\multirow{3}{*}{Top 7} & CS5K, CS15K & -0.02 & 0.08471 & 4 & -4.243 & 0.0286* \\
                       & CS5K, CS25K & -0.03 & 0.08471 & 4 & -6.364 & 0.0069** \\
                       & CS15K, CS25K & -0.01 & 0.08471 & 4 & -2.121 & 0.2006 \\
\bottomrule
\end{tabular}
\begin{tablenotes}[para,flushleft]
\item[*] \(p<.05\), ** \(p<.01\), CS: Customer Size
\end{tablenotes}
\end{threeparttable}
\end{table}

Table \ref{table:tukey_test} shows that for Top 1 and 3 in terms of CS, there is no significant difference between 5K and 15K, but there is a significant improvement from 15K to 25K. For Top 5, increasing CS from 5K to 15K or from 15K to 25K does not make a significant difference. Only increasing CS from 5K to 25K shows a significant improvement. For Top 7, increasing CS from 5K to 15K or 25K brings a significant improvement, while increasing it from 15K to 25K does not create a significant difference. Therefore, it can be recommended to prefer 25K for all top-N F1 scores.

\begin{figure}[H]
    \center
    \includegraphics[scale=0.9]{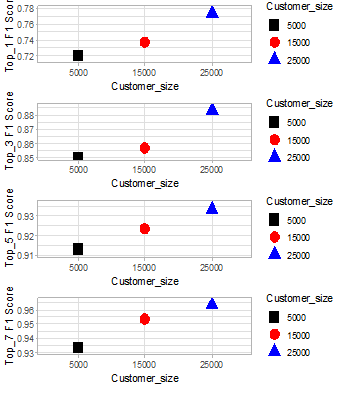}
    \caption{F1 Scores for the Customer Size}
    \label{fgr:maineffect}
\end{figure}

When Figure \ref{fgr:maineffect} is examined, it is observed that in all top 1, 3, 5, and 7 scores, the F1 scores increase as the customer size increases.

This analysis underscores the crucial role of customer size in enhancing the model's performance. On the other hand, the results from the factorial ANOVA also indicate that the window size does not have a significant impact on the performance of the model. It corroborates the notion that more extensive data leads to greater generalization and, consequently, improved performance, a principle particularly relevant for complex models like LSTM, as noted in prior studies by \cite{zhu2019electrocardiogram} and \cite{sun2020dl}.

\section{Conclusion}
\label{sec:conclusion}
Predicting the next destinations of customers has become a prominent and relevant topic in recent times. This is mostly because of all the advantages it provides businesses with, like maximising operational effectiveness, improving marketing strategies, and raising general customer satisfaction. Several strategies are used to accomplish this goal, however, developing a reliable model that can accurately predict the next destinations poses a significant challenge.

In this study, an LSTM model is developed to predict the next destinations of customers. The model is tested using a dataset obtained from an airline company. LSTM is a neural network model working very well with sequential data and has the capabilities to deal with higher complexity and long-term dependencies. To evaluate the impact of different factors on the results, the model is applied to a diverse range of customer populations and varying window sizes. By systematically varying the number of customers and window sizes, this study aims to uncover valuable insights into the influence of these variables on the performance of the model. Moreover, this empirical investigation provide valuable guidance for optimizing the model's configuration and enhancing its predictive accuracy in real-world scenarios. The results of the model indicate that the most favorable outcomes in terms of top-1, top-3, and top-7 F1 scores are obtained when the customer data is at its largest scale. It is observed that the performance of the model is positively influenced by the size of the customer data. However, the impact of the window size is not statistically significant and shows no effect on the performance. Overall, the obtained results give hope that the performance of the proposed method could also be high and satisfactory in other datasets.

There can be several reasons for not being able to reach high scores successfully. One of the major obstacles is the lack of data. If there is not enough historical data available on the travel patterns of customers, it can be challenging to build accurate prediction models. This is particularly true for new customers or customers who have not traveled frequently in the past. Another obstacle can be incomplete data, which is missing critical information such as the travel purpose, the customer's demographics, or the time of the year. This missing data can significantly impact the performance of the prediction models. Also, the complexity of travel behavior is of utmost importance in terms of building good models. Travel decisions of customers can be varied frequently depending on several different factors such as the purpose of travel, preferences, changes in the travel industry, economic conditions, etc. Another obstacle is unforeseen events, which can be called outliers as well. These conditions can be weather disruptions, political unrest, or health emergencies, which make it difficult to accurately predict the next destinations of customers. Taking all these circumstances into consideration, building robust models can be highly challenging.

Future studies could explore the use of more extensive datasets and more complex model architectures, as the analysis suggests that incorporating more data leads to improved performance. Another aspect that could be explored in future work is predicting the timing of the next travel. While this study focuses on predicting the next destination, forecasting the specific time at which a customer is likely to travel to that destination could provide valuable insights. It is worth considering that the structure of the dataset itself may have influenced the obtained results. Therefore, future research endeavors should delve deeper into exploring the potential impact of dataset characteristics on the performance of predictive models.

\bigskip
\RaggedRight
\textbf{Declaration of generative AI and AI-assisted technologies in the writing process}
\bigskip

During the preparation of this work the authors used ChatGPT in order to improve language and readability, with caution. After using this tool/service, the authors reviewed and edited the content as needed and take full responsibility for the content of the publication.

\bibliographystyle{elsarticle-harv} 
\bibliography{references} 





\end{document}